\documentclass[letterpaper, 10 pt, conference]{ieeeconf}  

\IEEEoverridecommandlockouts                              

\usepackage{graphics} 
\usepackage{epsfig} 
\usepackage{times} 
\usepackage{amsmath} 
\usepackage{amssymb}  
\usepackage{biblatex}
\usepackage{multirow}
\usepackage{multicol}
\usepackage{hyperref}
\usepackage{booktabs}
\usepackage{enumerate}
\usepackage{color}
\usepackage{tabularx}
\usepackage{hyperref}
\usepackage{pifont}
\usepackage{algorithm}
\usepackage{algpseudocode}
\usepackage{amsmath}
\usepackage{adjustbox}
\usepackage[usenames,dvipsnames,table]{xcolor}
\hypersetup{
    colorlinks=true,
    linkcolor=MidnightBlue,
    filecolor=magenta,      
    urlcolor=MidnightBlue,
    citecolor=MidnightBlue,
} 
\usepackage[font=small,labelfont=bf]{caption}

\definecolor{darkred}{rgb}{0.76, 0.23, 0.13}
\definecolor{darkgreen}{rgb}{0.01, 0.75, 0.24}
\definecolor{darkgray}{rgb}{0.66, 0.66, 0.66}

\title{\LARGE \bf
Towards Effective Utilization of Mixed-Quality Demonstrations in Robotic Manipulation via Segment-Level Selection and Optimization
}

\author{Jingjing Chen$^{1}$, Hongjie Fang$^{1}$, Hao-Shu Fang$^{1}$ and Cewu Lu$^{1,2,\dagger}$
\thanks{$^{1}$Shanghai Jiao Tong University.}
\thanks{$^{2}$Shanghai Innovation Institute.}
\thanks{$^\dagger$ Cewu Lu is the corresponding author.}
\thanks{\{jjchen20, galaxies, lucewu\}@sjtu.edu.cn, fhaoshu@gmail.com.}
}

\begin{document}

\maketitle
\thispagestyle{empty}
\pagestyle{empty}

\begin{abstract}

Data is crucial for robotic manipulation, as it underpins the development of robotic systems for complex tasks. While high-quality, diverse datasets enhance the performance and adaptability of robotic manipulation policies, collecting extensive expert-level data is resource-intensive. Consequently, many current datasets suffer from quality inconsistencies due to operator variability, highlighting the need for methods to utilize mixed-quality data effectively.
To mitigate these issues, we propose ``\textit{Select Segments to Imitate}'' (S2I), a framework that selects and optimizes mixed-quality demonstration data at the segment level, while ensuring plug-and-play compatibility with existing robotic manipulation policies. The framework has three components: demonstration segmentation dividing origin data into meaningful segments, segment selection using contrastive learning to find high-quality segments, and trajectory optimization to refine suboptimal segments for better policy learning.
We evaluate S2I through comprehensive experiments in simulation and real-world environments across six tasks, demonstrating that with only 3 expert demonstrations for reference, S2I can improve the performance of various downstream policies when trained with mixed-quality demonstrations. Project website: \href{https://tonyfang.net/s2i/}{https://tonyfang.net/s2i/}.
\end{abstract}

\section{Introduction}\label{sec:introduction}

In the realm of robot manipulation, data serves as the cornerstone for developing effective and adaptive systems~\cite{rosie}. Robots rely on extensive datasets~\cite{roboset, oxe, rh20t, droid, bridgedatav2} to acquire the necessary knowledge and skills for performing complex tasks. The quality, diversity, and quantity of demonstration data directly affect the accuracy and generalization ability of robotic manipulation policies. Without sufficient and well-curated data, robots may struggle to achieve reliable performance and adaptability. Therefore, data is not merely a supplementary component but a fundamental element in the advancement and success of robot manipulation policies~\cite{rt1}.

While demonstration data is essential for robotic manipulation, it often varies in quality and consistency due to differences in the expertise of data collection personnel~\cite{robomimic}.
For example, when collecting robot demonstration data using teleoperation devices~\cite{ding2024bunny, airexo, act}, less experienced operators may make errors or perform unnecessary actions, leading to a decrease in the quality of the demonstration data. 
Although high-quality expert demonstrations are ideal for robotic manipulation policy training, acquiring a large amount of such data is both difficult and resource-intensive~\cite{rt1,rh20t}. 

As a result, these challenges underscore the need for methods that effectively utilize and enhance available mixed-quality demonstration data, especially when policies must \textit{rely solely on existing data} for learning, without the option to add new data~\cite{ross2011reduction} or interact with the environment~\cite{gail}. Moreover, the method should aim to \textit{use as few manual annotations or expert demonstrations as possible} to reduce the labor cost of data processing and optimization.

There are two main lines of methods to address the issue: (1) improving the policy to handle mixed-quality data, often through weighted behavior cloning~\cite{bcnd}, which requires extensive labeled data for better performance~\cite{ileed, swbt, dwbc, cail}, and (2) optimizing the data to focus on higher-quality demonstrations. The latter can involve filtering at the demonstration level~\cite{elicit, l2d}, potentially missing high-quality segments, or at the state-action pair level~\cite{pubc}, which may overlook contextual relevance. Many existing methods simply discard low-quality data without optimizing them, leading to ineffective utilization of mixed-quality datasets.

\begin{figure}[t]
    \centering
    \includegraphics[width=1\linewidth]{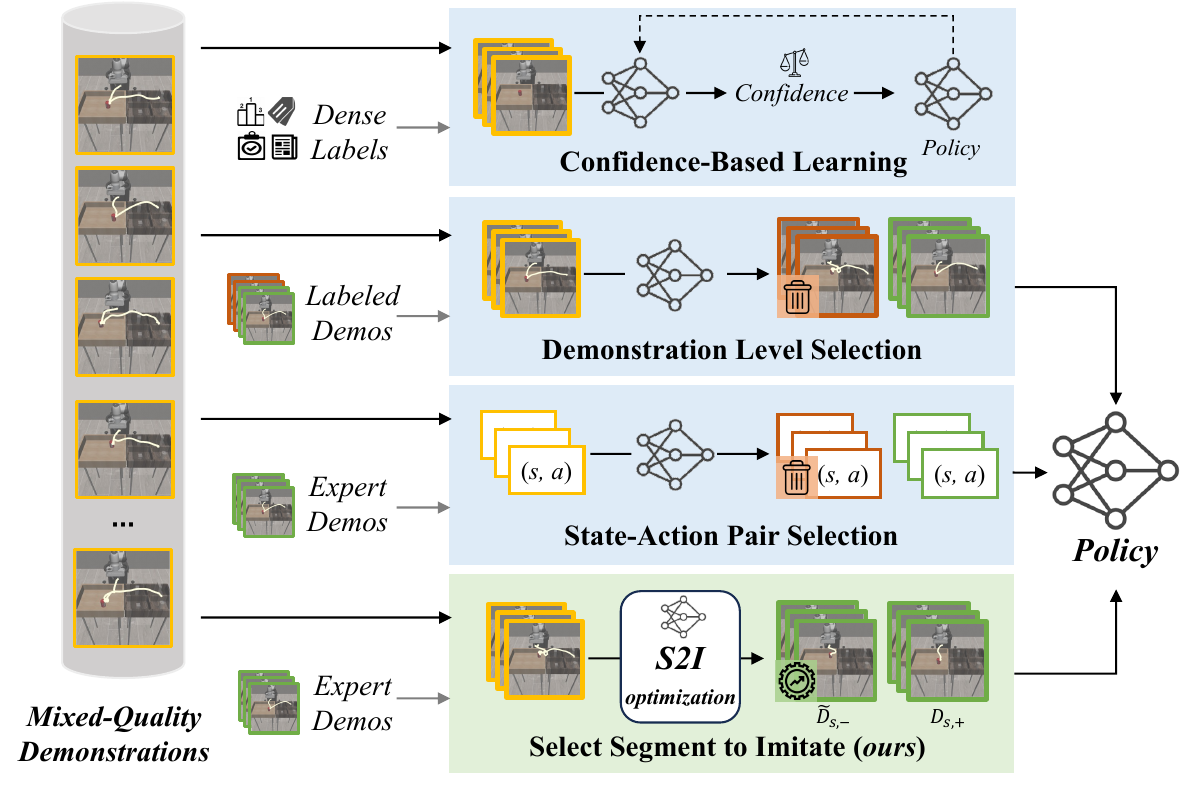}
    \caption{\textbf{Methods to Deal with Mixed-Quality Demonstrations}. Compared with previous methods, our S2I framework selects and optimizes demonstrations at the segment level, preserving semantic consistency within segments while maximizing the retention of high-quality segments and improving lower-quality ones, leading to effective utilization of the mixed-quality demonstrations. With only 3 expert demonstrations for reference, the plug-and-play S2I framework can be used as a data preprocessing step to enhance the performance of various downstream robot manipulation policies when handling mixed-quality demonstrations.}
    \label{fig:intro}
    \vspace{-0.5cm}
\end{figure}

To this end, we propose ``\textit{Select Segments to Imitate}'' (S2I), a framework that enhances the performance of robotic manipulation policies through effective utilization of the mixed-quality demonstrations, with the assistance of only 3 expert demonstrations. After dividing demonstrations into semantically consistent segments, our S2I framework processes at the segment level to extract more high-quality demonstration segments while preserving contextual information. To make the most of the available demonstrations, we also perform trajectory optimization and action relabeling on low-quality segments, which allows the policy to learn smooth trajectories and corrective behaviors. Experiments on both simulation environments and the real-world platform demonstrate the effectiveness of our S2I framework in utilizing mixed-quality demonstrations.

\section{Related Works}\label{sec:related-works}

\subsection{Learning from Mixed-Quality Demonstrations}

Imitation learning from mixed-quality demonstrations aims to learn a policy from demonstrations that may vary greatly in quality, and it can be roughly divided into two categories: confidence learning and demonstration selection.

\paragraph{Confidence Learning}
Several studies address the challenge of learning from mixed-quality demonstrations by introducing confidence weights to assess the optimality of state-action pairs~\cite{ileed, demodice, bcnd, swbt, wu2019imitation, dwbc, cail}. BCND~\cite{bcnd} uses the policy from previous steps as weights for iterative refinement. DWBC~\cite{dwbc} and ILEED~\cite{ileed} enhance policy performance through joint training with additional components: a discriminator and a state-encoder, respectively. However, these methods risk guiding the model towards incorrect distributions or increasing sensitivity to hyperparameters and training instability. Concurrently, a recent approach~\cite{yin2024offline} constructs a state graph from the mixed-quality demonstrations, linking nearby states in visual representation space and consecutive states in demonstrations to determine confidence through graph search. Still, the effectiveness depends on carefully tuning the tolerance range for nearby states.

\paragraph{Demonstration Selection}

The demonstration selection method decouples data processing from policy learning, simplifying the handling of mixed-quality data.
L2D~\cite{l2d} employs preference learning to filter out low-quality demonstrations, while ELICIT~\cite{elicit} aligns demonstrations with expert policy but may neglect the usefulness of low-quality data. PUBC~\cite{pubc} creates negative samples by mismatching state-action pairs and updates the positive dataset iteratively through adaptive voting, , overlooking the significance of sequences. AWE~\cite{awe} applies dynamic programming to smooth expert trajectories, potentially preserving outliers in mixed-quality data. These methods reflect the difficulty of utilizing mixed-quality demonstrations without oversimplifying the data. Moreover, simply discarding the low-quality data~\cite{bu2024aligning, elicit, l2d, pubc} can lead to inadequate  data utilization and limit downstream policy performance.

\subsection{Robotic Manipulation Policies}

Keyframe control~\cite{polarnet, act3d, rvt, arm, c2farm, peract, same} and continuous control~\cite{dp, robomimic, dexcap, rise, dp3, act} are two commonly used control strategies in robotic manipulation policies. The former only predicts keyframe actions, while the latter forecasts continuous action trajectories. Continuous control is more general than keyframe control and can be applied to a broader range of robotic manipulation scenarios~\cite{sgrv2}. Thus, we mainly focus on improving its performance in this paper. 

Based on the observation modality, the continuous manipulation policies can be roughly divided into three categories: 1D state-based~\cite{dp, robomimic, bet}, 2D image-based~\cite{dp, robomimic, act}, and 3D point-cloud-based~\cite{dexcap, rise, dp3, sgrv2}. Among these approaches, BC-RNN~\cite{robomimic} integrates temporal information to improve behavior cloning~\cite{bc}, Diffusion Policy (DP)~\cite{dp} models action prediction as the diffusion denoising process~\cite{diffusion_model}, and ACT~\cite{act} trains a Transformer-based policy using a CVAE~\cite{sohn2015learning} scheme. 3D-based methods typically employ point-based encoders~\cite{pointnet, dexcap, dp3, sgrv2} or sparse convolutional encoders~\cite{minkowski, rise} to perceive point cloud inputs.

\subsection{Contrastive Learning}
Contrastive learning has made significant strides in representation learning~\cite{caron2020unsupervised, simclr, byol, moco, supcon}.
In robotics, contrastive learning has been applied in various domains such as visual representation learning~\cite{vip, r3m}, multimodal representation learning~\cite{liv, actra}, and offline data retrieval~\cite{du2023behavior, vinn}. L2D~\cite{l2d} also utilizes contrastive learning to effectively extract behavior features at the demonstration level, showcasing the versatility of this approach across mixed-quality demonstrations. This broad applicability highlights the potential of contrastive learning to unify learning strategies across different data modalities and improve overall task performance.

\section{Select Segments to Imitate}\label{sec:s2i}

\begin{figure*}
    \centering
    \includegraphics[width=0.9\linewidth]{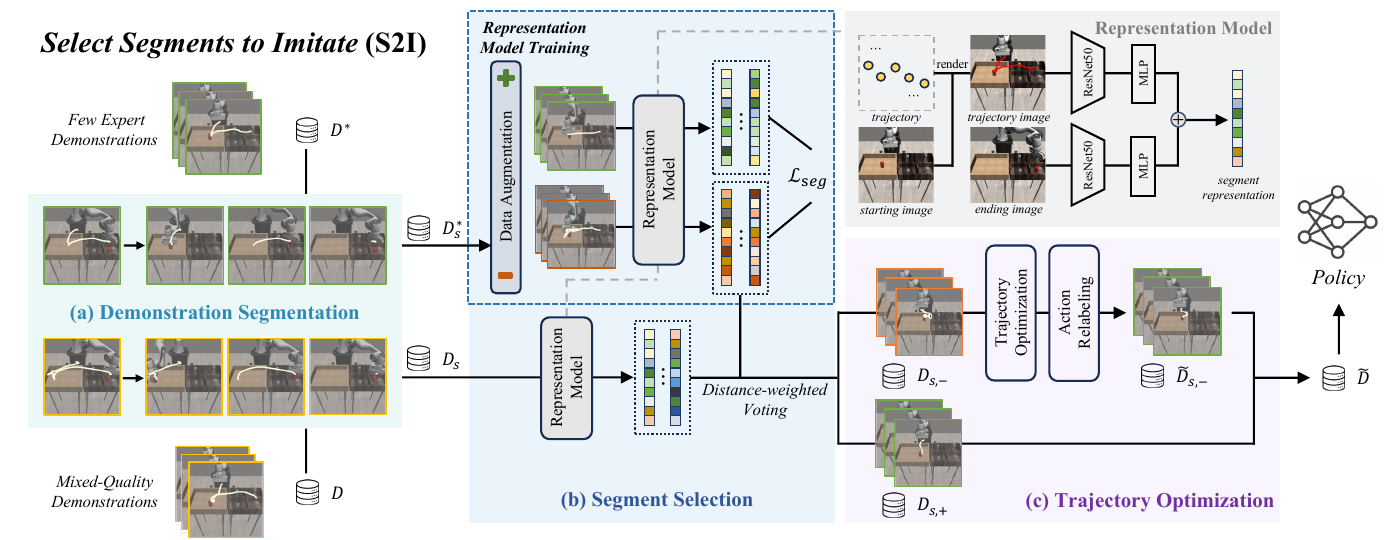}
    \caption{\textbf{Overview of the \textit{Select Segments to Imitate} (S2I) Framework.} (a) \textit{Demonstration segmentation stage} divides demonstrations into semantic-meaningful segments; (b) \textit{Segment selection stage} applies contrastive learning to train the representation model on the expert demonstration segments, and then use distance-weighted voting to determine the quality of the mixed-quality demonstration segments; (c) \textit{Trajectory optimization stage} optimizes the robot trajectory within the low-quality segments and perform action relabeling for the efficient utilization of the whole demonstration dataset (for details, please refer to Fig.~\ref{fig:traj-opt}). Finally, the high-quality segments and the optimized low-quality segments form the final optimized dataset $\tilde{\mathcal{D}}$, which can be used directly for downstream policy learning.}
    \label{fig:s2i} \vspace{-0.4cm}
\end{figure*}

In this section, we first formally describe the problem of learning from mixed-quality demonstration data within the context of robotic manipulation (\S \ref{sec:problem_formulation}). 
We then propose ``\textit{Select Segments to Imitate}'' (S2I), a framework for processing mixed-quality datasets, which consists of three parts: demonstration segmentation (\S \ref{sec:segment_demonstration}), segment selection (\S \ref{sec:segment_selection}), and trajectory optimization (\S \ref{sec:trajectory_optimization}). The overview of our S2I framework is illustrated in Fig.~\ref{fig:s2i}.

\subsection{Problem Formulation}\label{sec:problem_formulation}
Consider an offline dataset $\mathcal{D} = \{\tau_1, \tau_2, \ldots, \tau_N\}$ containing $N$ mixed-quality demonstrations for a robotic manipulation task. Each demonstration $\tau$ is a trajectory $\left(o_1, a_1, o_2, a_2, \ldots, o_T, a_T\right)$, where $t \in \{1, 2, \ldots, T\}$ represents the timestep, $o_t$ is the observation, and $a_t$ is the action taken by the robot. Additionally, a small offline dataset $\mathcal{D}^*=\{\tau^*_1,\tau^*_2,\cdots,\tau^*_M\}$ of $M \ll N$ expert demonstrations is also provided for reference and cannot be used for policy learning. In some cases, each demonstration $\tau_i$ in $\mathcal{D}$ may have a subjective quality label $l_i$, which is assumed to be inaccessible in our problem setting.

The goal is to learn a robotic manipulation policy from the mixed-quality dataset $\mathcal{D}$, leveraging the small expert dataset $\mathcal{D}^*$, with no additional data collection or environmental interaction allowed during the learning process.

\subsection{Demonstration Segmentation}\label{sec:segment_demonstration}

Segmenting demonstrations helps identify high-quality segments, particularly within low-quality data. In long-horizon tasks with multiple subtasks, human teleoperated demonstrations may exhibit instability in some subtasks but accuracy in others. By dividing a demonstration $\tau$ into segments $(\tau^{(1)}, \tau^{(2)}, \cdots)$, we can retain high-quality segments for policy learning while optimizing or discarding lower-quality ones. Accurate detection of transition points between segments is crucial: well-chosen points ensure meaningful segments, while poor choices may result in incoherent segments. Thus, a robust segmentation method is required for handling low-quality demonstrations effectively.

\begin{figure}[t]
    \centering
    \includegraphics[width=\linewidth]{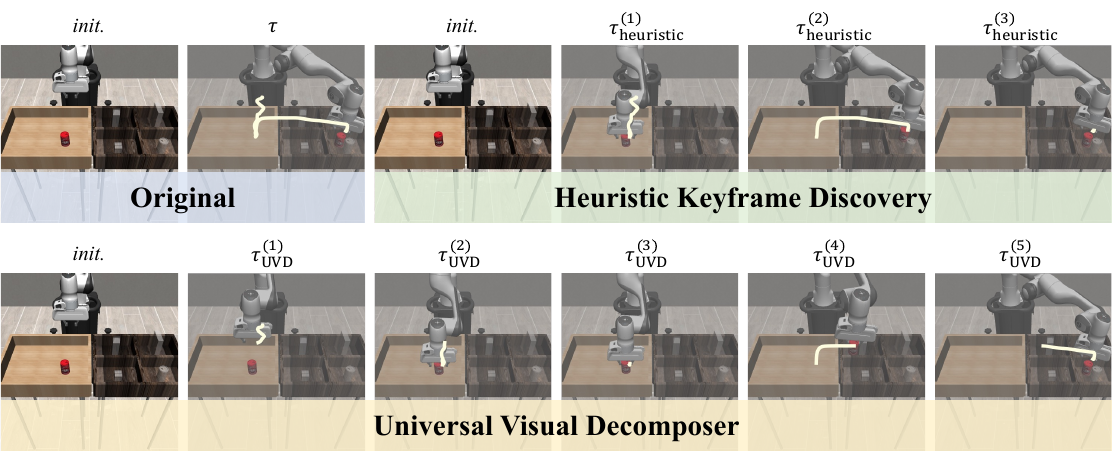}
    \caption{\textbf{Segmentation Results of a RoboMimic-\textit{Can}~\cite{robomimic} Demonstration}. UVD might produce meaningless and inconsistent segments like $\tau_\text{UVD}^{(3)}$ and $\tau_\text{UVD}^{(4)}$ compared to heuristic keyframe discovery.}
    \label{fig:seg-result}\vspace{-0.5cm}
\end{figure}

We consider two candidates for demonstration segmentation: \textit{heuristic keyframe discovery}~\cite{arm,c2farm,peract} and \textit{universal visual decomposer (UVD)}~\cite{uvd}. Heuristic keyframe discovery identifies a frame as a keyframe based on either a change in gripper state or robot velocities approaching zero. In contrast, UVD uses pre-trained visual representations~\cite{liv, vip, r3m, dinov2, clip} to extract image features and recursively searches for keyframes through distance measurements. These keyframes are then used to divide the demonstration into segments. After visualizing the segmentation results in Fig.~\ref{fig:seg-result}, we observe that visual changes can heavily affect the performance of UVD, leading to excessive and meaningless segments in low-quality demonstrations while heuristic keyframe discovery maintains complete and accurate segmentation for both demonstrations. Hence, we choose heuristic keyframe discovery to ensure effective and robust demonstration segmentation. After segmentation, we obtain mixed-quality demonstration segment dataset $\mathcal{D}_s$ and expert demonstration segment dataset $\mathcal{D}_s^*$. In the following sections, we omit the superscript and use $\tau$ to denote the trajectory segment for brevity.

\subsection{Segment Selection}\label{sec:segment_selection}

After demonstration segmentation, we need to select high-quality segments from $\mathcal{D}_s$ with the assistance of $\mathcal{D}_s^*$. To achieve this, we employ supervised contrastive learning~\cite{supcon} for segment representation learning.

To assess the quality of a demonstration segment, we consider two key factors: (1) the starting and ending states, which indicate action success or failure, and (2) the robot trajectory, which reflects task execution quality. We use these factors by inputting the starting and ending images of the segment, along with the robot trajectory, into a segment representation learning process. Given the multimodal nature of the inputs, which include both low-dimensional trajectories and images, effective fusion or alignment of these features is essential. Inspired by \cite{vosylius2024render}, we align the robot trajectory with the images by rendering the trajectory onto the starting image. We then pass the starting image with the rendered trajectory and the ending image through ResNet-50 encoders~\cite{resnet} for feature extraction. The extracted features are separately projected and concatenated to create the segment representation $\mathbf{z}$.

Given that the expert demonstration segment dataset $\mathcal{D}^*_s$ is too small, we perform data augmentations to generate positive and negative samples based on the dataset. Positive augmentation creates trajectory images from various perspectives of the original demonstration, while negative augmentation either introduces noise to trajectories or mismatches the ending image with the starting image with the rendered trajectory.
After that, we obtain augmented positive segment dataset $\mathcal{D}^*_s\cup \mathcal{D}^*_{s,+}$ and negative segment dataset $\mathcal{D}^*_{s,-}$. The following objective is applied for segment representation learning~\cite{supcon}.
$$
\begin{aligned}
    \mathcal{L}_\text{seg} = &-\frac{1}{\left|\mathcal{B}_+\right|} \sum_{a\in\mathcal{B}_+} \sum_{b\in\mathcal{B}_+\setminus \{a\}} \log \frac{\exp(\textbf{z}_a \cdot \textbf{z}_b / t)}{\sum_{b'\in\mathcal{B}\setminus\{a\}} \exp(\textbf{z}_a \cdot \textbf{z}_{b'} / t)} \\ &-\frac{1}{\left|\mathcal{B}_-\right|} \sum_{a\in\mathcal{B}_-} \sum_{b\in\mathcal{B}_-\setminus \{a\}} \log \frac{\exp(\textbf{z}_a \cdot \textbf{z}_b / t)}{\sum_{b'\in\mathcal{B}\setminus\{a\}} \exp(\textbf{z}_a \cdot \textbf{z}_{b'} / t)} 
\end{aligned}
$$
where $\mathcal{B}$ denotes a batch of segments during training, $\mathcal{B}_+ := \mathcal{B} \cap (\mathcal{D}_s^* \cup \mathcal{D}_{s,+}^*)$ is the set of positive segments in the batch, $\mathcal{B}_- := \mathcal{B}\cap\mathcal{D}_{s,-}^*$ is the set of negative segments in the batch, and $t$ is the temperature parameter.

After training the segment representation model, we extract the segment representation $\mathbf{z}_{\tau}$ for all $\tau \in \mathcal{D}_s$. Then the quality label of $\tau$ is determined via distance-weighted voting~\cite{dudani1976distance} from labeled dataset $\mathcal{D}_s^*\cup\mathcal{D}_{s,+}^*\cup\mathcal{D}_{s,-}^*$, \textit{i.e.}, $\tau$ is classified as a high-quality segment if and only if
$$
\frac{\sum_{a\in N_{k,+}(\tau)} \exp\left(-\left\|\textbf{z}_\tau - \textbf{z}_a\right\|\right)}{\sum_{a\in N_k(\tau)} \exp\left(-\left\|\textbf{z}_\tau - \textbf{z}_a\right\|\right)} \ge \delta_c
$$
where $N_k(\tau)$ is the $k$-nearest neighbors of $\tau$ in the labeled dataset, $N_{k,+}(\tau) := N_k(\tau) \cap (\mathcal{D}_s^*\cup\mathcal{D}_{s,+}^*)$ is the positive-labeled trajectory set in $N_k(\tau)$, and $\delta_c$ is the threshold for classification. By labeling all segments in $\mathcal{D}_s$, we can create positive segment dataset $\mathcal{D}_{s,+}$ directly for policy training and negative segment dataset $\mathcal{D}_{s,-}$ for trajectory optimization.

\subsection{Trajectory Optimization}\label{sec:trajectory_optimization}

Reducing errors is essential to fully utilize the valuable information in the negative segment dataset $\mathcal{D}_{s,-}$. By focusing on rearranging states closer to the correct path, the downstream policy can learn more accurate action patterns.
Meanwhile, models often struggle when faced with unseen states in imitation learning, as they have difficulties in inferring the correct behavior in out-of-distribution scenarios. Hence, it is necessary to recompute the actions for the states discarded in trajectory optimization, ensuring the policy can make full use of dataset by learning from relabeled actions.

Since the demonstration segment $\tau$ contains semantically consistent robot actions, we propose a greedy algorithm (Alg.~\ref{alg:greedy_trajectory_optimization}) to approximate the optimal path $\tau'$ by eliminating points with excessive spatial deviation. For $\tau \in \mathcal{D}_{s,-}$, we start with $\tau' = \{(o_1, a_1)\}$ and iteratively select the next waypoint. Define $\mathbf{e}_{i,j} := e_i - e_j$ as the vector from $e_j$ to $e_i$, where $e_i$ represents the end-effector position at point $i$. We construct a candidate set $K$ for the next waypoint $(o_k, a_k)$ using a threshold $\delta_\theta$ for the maximum angle between $\mathbf{e}_{k,j}$ and $\mathbf{e}_{T,j}$, where $T$ is the ending timestep of the segment. This process filters out unnecessary actions that deviate significantly from the action trend within the segment. We then choose $(o_k, a_k)$ with the minimum distance $\left|\mathbf{e}_{k,j}\right|$ as the next waypoint and add it to $\tau'$. The selection process continues until $(o_T, a_T)$ is included, forming the approximated optimal path $\tau'$ for the low-quality segment $\tau$.

\begin{figure}[t]
\vspace{-0.25cm}
    \centering
    \includegraphics[width=0.9\linewidth]{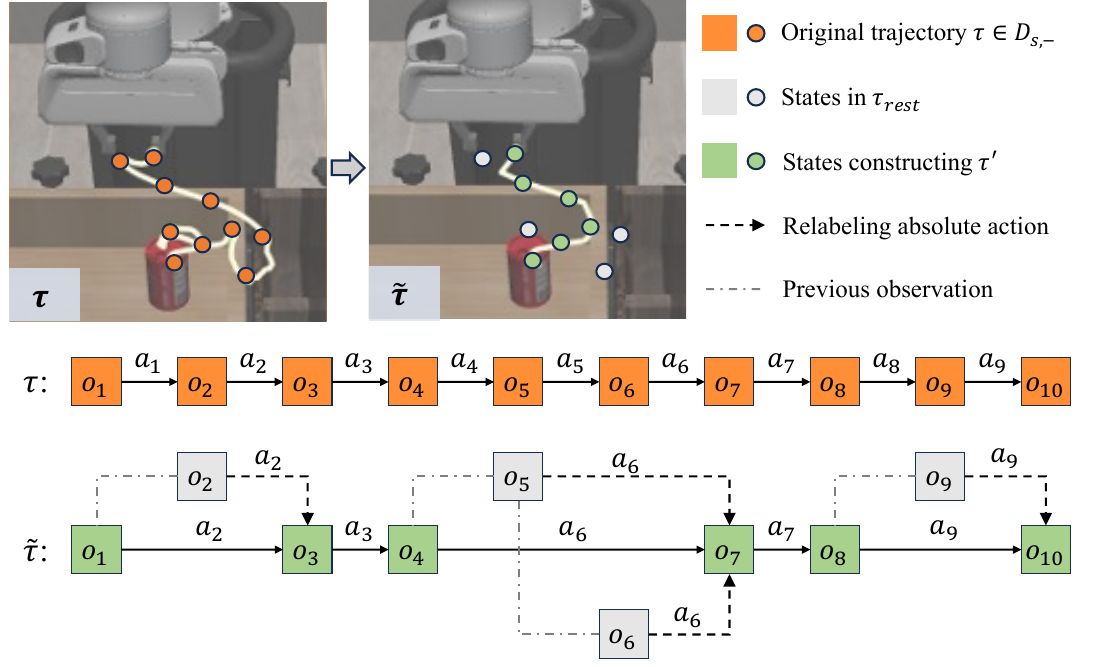}
    \caption{\textbf{Trajectory Optimization and Action Relabeling}. Orange points stand for the original trajectory, green ones denote optimized trajectory $\tau'$, and grey ones are discarded points. We perform action relabeling on all points in the original trajectory $\tau$ to form the optimized ``trajectory'' $\tilde{\tau}$. The example here assumes absolute actions, but S2I can support both absolute and relative actions.}
    \label{fig:traj-opt}\vspace{-0.5cm}
\end{figure}

After applying the greedy algorithm, we can use action relabeling on the discarded waypoints to preserve as much demonstration data as possible. By assuming absolute actions, the action relabeling can be written as $\tilde{a}_{t} = a_{t'}$ where $ t' = \min\{  t' \mid  t' \ge t \text{ and } \tilde{a}_{t'+1} \in \tau' \}$, as illustrated in Fig.~\ref{fig:traj-opt}. Thus, the optimized negative dataset $\tilde{\mathcal{D}}_{s,-}$ comprises the original trajectory $\tau$ with the relabeled actions $\{(o_t,\tilde{a}_t)\}_{t=1}^{T}$ derived from the optimized trajectory $\tau'$, which we denote as $\tilde{\tau}$. This enables $\tilde{\mathcal{D}}_{s,-}$ to support manipulation policies with historical observation horizon~\cite{dp} or action chunking~\cite{act}. Finally, we obtain the optimized dataset $\tilde{\mathcal{D}} = \mathcal{D}_{s,+}\cup \tilde{\mathcal{D}}_{s,-}$ for downstream policy learning.

\begin{algorithm}[t]
\small
\caption{\small Greedy Algorithm for Trajectory Optimization}\label{alg:greedy_trajectory_optimization}

\textbf{Input:} $\tau = \left(o_1, a_1,  \cdots, o_{T}, a_{T}\right) \in \mathcal{D}_{s,-}$, tolerance angle $\delta_\theta$.

\textbf{Output:} Optimized segment $\tau'$ of negative segment $\tau$.

\begin{algorithmic}[1]
\State Initialize $\tau' \gets \{(o_1, a_1)\}$, $\tau_\text{rest} \gets \{(o_i, a_i)\}_{i=2}^{T}$.
\State $\delta_s \gets \max_{i=1}^{T-1} \|\mathbf{e}_{i+1,i}\| $.

\While{$(o_T, a_T) \notin \tau'$}
    \State Let $(o_j, a_j)$  be the last element in $\tau'$.
    \State $K \gets \{ k \mid (o_k,a_k)\in \tau_\text{rest}, \angle(\mathbf{e}_{k,j}, \mathbf{e}_{T,j}) \le \delta_\theta \}$.
    \If {$K$ is $\varnothing$}
        \State $K \gets \{ k \mid (o_k,a_k)\in \tau_\text{rest}, \|\mathbf{e}_{k,j}\| \geq \delta_s\}$.
    \EndIf
    \State $k\gets \arg\min_{k \in K} \|\mathbf{e}_{k,j}\|$.
    \State Update $\tau' \gets \tau' \cup \{(o_k, a_k)\}$.
    \State Update $\tau_\text{rest} \gets \tau_\text{rest} \setminus \{(o_k, a_k)\}$.
\EndWhile
\State \textbf{Return} $\tau'$.
\end{algorithmic}
\end{algorithm}

\section{Experiments}\label{sec:experiments}

In this section, we aim to answer the following questions: \textbf{(Q1)} Can S2I improve the performance of robotic manipulation policies (\S \ref{sec:sim})? \textbf{(Q2)} What types of data are beneficial for policy learning (\S \ref{sec:sim})? \textbf{(Q3)} How do different design choices affect the performance of S2I (\S \ref{sec:ablation})? \textbf{(Q4)} How effective is S2I in processing real-world demonstration data and improving downstream policy learning (\S \ref{sec:real-world})?

\begin{figure*}
    \centering
    \includegraphics[width=0.75\linewidth]{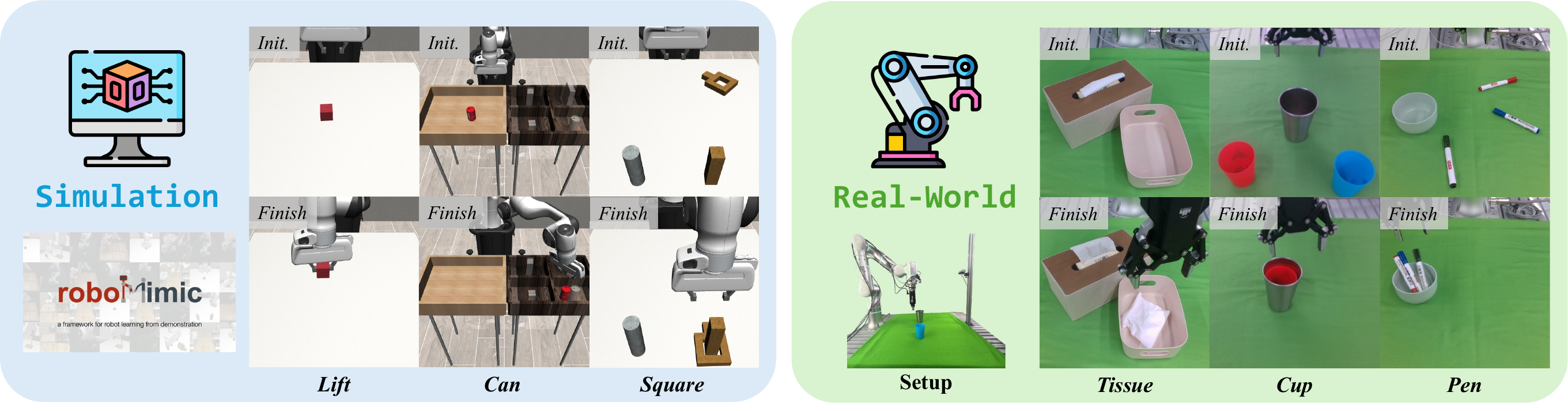}
    \vspace{-0.1cm}
    \label{fig:tasks}
    \caption{\textbf{Tasks}. We select 3 tasks from RoboMimic~\cite{robomimic} (\textbf{\textit{Lift}}, \textbf{\textit{Can}} and \textbf{\textit{Square}}) in the simulation environment, and design 3 tasks (\textbf{\textit{Tissue}}, \textbf{\textit{Cup}} and \textbf{\textit{Pen}}) in the real-world robot platform for evaluation. Detailed descriptions of all tasks: (1) \textbf{\textit{Lift}}: pick the block; (2) \textbf{\textit{Can}}: move the can into the target area; (3) \textbf{\textit{Square}}: pick a square nut and place it on the target rod; (4) \textbf{\textit{Tissue}}: take tissue out and place it in the container; (5) \textbf{\textit{Cup}}: collect all the cups (at most 2) into the large metal cup; (6) \textbf{\textit{Pen}}: collect all the pens (at most 3) into the bowl. }
    \label{fig:tasks}\vspace{-0.35cm}
\end{figure*}

\begin{figure*}
\begin{adjustbox}{valign=t,minipage={0.65\linewidth}}
    \centering
    \includegraphics[width=\linewidth]{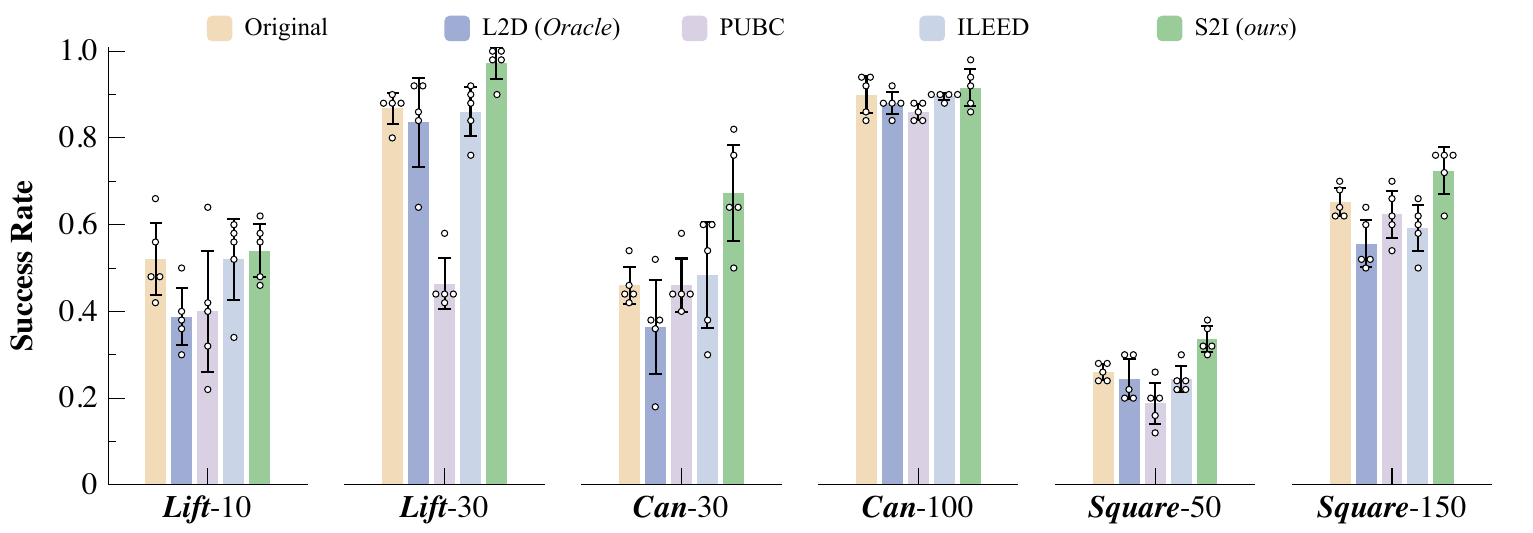}
    \captionof{figure}{\textbf{BC-RNN Performance on the State-Based RoboMimic Benchmark.} The success rates are averaged across 5 different seeds, based on the checkpoint with the best performance. Our S2I framework effectively and consistently enhances success rates across different tasks with different amounts of demonstration data, outperforming baselines. }\label{fig:result}
\end{adjustbox}
\hfill
\begin{adjustbox}{valign=t,minipage={0.33\linewidth}}
    \centering
    \footnotesize
    \renewcommand{\arraystretch}{1.1}
    \centering
    \setlength\tabcolsep{2pt}
    \begin{tabular}{ccr}
    \toprule
    \textbf{Method} & \begin{tabular}{c}\textbf{Dataset}\\ \textbf{Utilization}\end{tabular} & \begin{tabular}{c}\textbf{Average}\\ \textbf{Gain $\uparrow$}\end{tabular} \\
    \midrule
    Original & 100.00\% & {\color{darkgray}$0.00\%$} \\
    \midrule
    L2D \cite{l2d} (\textit{Oracle}) & 51.59\% & {\color{darkgreen}$-12.37\%$} \\
    PUBC \cite{pubc} & 89.49\% & {\color{darkgreen}$-18.58\%$} \\
    ILEED \cite{ileed} & 100.00\% & {\color{darkred}$+0.97\%$} \\
    \midrule
    S2I (ours) & 100.00\% & {\color{darkred}$\mathbf{+21.16\%}$} \\
    \bottomrule
    \end{tabular}
    \captionof{table}{\textbf{BC-RNN Performance on the State-Based RoboMimic Benchmark.} The results are averaged across 5 different seeds, based on the last 10 checkpoints. The performance gain is calculated \textit{w.r.t.} original BC-RNN and then averaged over 6 experiments.}\label{tab:stability}
\end{adjustbox}
\vspace{-0.6cm}
\end{figure*}

\subsection{Setup}

\paragraph{Tasks} For simulation experiments, we evaluate three RoboMimic~\cite{robomimic} tasks (\textbf{\textit{Lift}}, \textbf{\textit{Can}}, \textbf{\textit{Square}}) in RoboSuite~\cite{robosuite}, and for real-world experiments, we evaluate three tasks (\textbf{\textit{Tissue}}, \textbf{\textit{Cup}}, \textbf{\textit{Pen}}) on a platform with a Flexiv Rizon robotic arm, Dahuan AG-95 gripper and 2 Intel RealSense D435 cameras. Task details are shown in Fig.~\ref{fig:tasks}.

\paragraph{Mixed-Quality Demonstration Dataset} 
We use the RoboMimic multi-human (MH) dataset~\cite{robomimic} for simulation experiments, which includes demonstration data of varying quality from 6 operators. We sample multiple demonstrations per task from the dataset, ensuring a balanced mix of demonstration qualities, and use \textbf{\textit{A}}-$N$ to denote the results obtained from using $N$ mixed-quality demonstrations in task \textbf{\textit{A}}. For real-world experiments, we collect 25 expert and 25 suboptimal demonstrations per task, combining them into 50 mixed-quality datasets. For reference expert dataset, we select $M=3$ demonstrations.

\paragraph{Robotic Manipulation Policies} For simulation experiments, we use the 1D state-based BC-RNN~\cite{robomimic} and the Diffusion Policy (DP)~\cite{dp} that can be applied to both 1D state and 2D image data as robot manipulation policies. For real-world experiments, we choose DP and ACT~\cite{act} as our 2D image-based policies, as well as RISE~\cite{rise} as our 3D point-cloud-based policy. The hyper-parameters of all policies remain unchanged.

\paragraph{Baselines} For RoboMimic simulation experiments, we choose ILEED~\cite{ileed} for confidence-based learning, L2D~\cite{l2d} for demonstration-level selection, 
and PUBC~\cite{pubc} for state-action pair selection as baseline methods. Since L2D is not open-sourced but has reported high demonstration-level classification accuracy with ground-truth labels, we use ``better'' and ``okay'' labeled demonstrations for downstream policy learning, denoted as ``L2D (\textit{Oracle})''.

\paragraph{Metrics} We use the success (completion) rate as the main metric. Performance gain $\frac{\text{Success Rate} - \text{Original Success Rate}}{\text{Original Success Rate}}$ is also reported to evaluate the relative improvements.

\begin{table*}
\renewcommand{\arraystretch}{1.1}
    \centering
    \begin{tabular}{c|ccc|ccc}
    \toprule
    \multirow{2}{*}{\textbf{Method}} & \multicolumn{3}{c|}{\textbf{State-Based}} & \multicolumn{3}{c}{\textbf{Image-Based}} \\ 
    & \textbf{\textit{Lift}}-10 & \textbf{\textit{Can}}-30 & \textbf{\textit{Square}}-50 & \textbf{\textit{Lift}}-10 & \textbf{\textit{Can}}-30 & \textbf{\textit{Square}}-50 \\
    \midrule
    Original &  0.952 / \textbf{0.912}  & 0.824 / 0.749  & 0.504 / 0.409  & \textbf{0.964} / 0.743  & 0.640 / 0.547 & 0.568 / 0.492 \\
    \midrule
    L2D~\cite{l2d} (\textit{Oracle}) & 0.544 / 0.457  & 0.796 / 0.734  & 0.368 / 0.258  & 0.708 / 0.442 & 0.524 / 0.439 & 0.444 / 0.377 \\
    PUBC~\cite{pubc} & 0.792 / 0.723 & 0.756 / 0.680 & 0.316 / 0.250 & - & - & -\\
    \midrule
    S2I (\textit{ours}) & \textbf{ 0.964} / 0.906 & \textbf{0.840} / \textbf{0.785} & \textbf{0.580} / \textbf{0.500} & 0.956 / \textbf{0.867} & \textbf{0.712} / \textbf{0.624} & \textbf{0.604} / \textbf{0.506}\\
    \bottomrule
    \end{tabular}
    \caption{\textbf{DP Performance on the RoboMimic Benchmark\protect\footnotemark.} The success rates are averaged across 5 different seeds. Following \cite{dp}, the same format of (best performance) / (average of last 10 checkpoints) is reported here.}\label{tab:res-dp}\vspace{-0.55cm}
\end{table*}

\paragraph{Protocols} Due to variations in data scale from different data processing methods, we train the policy for a fixed number of iterations, each with a randomly sampled batch from the processed data. During the simulation evaluation, we randomly select 50 starting positions, which vary across seeds but are consistent with the same seed. For real-world experiments, we perform 20 consecutive trials with randomly placed objects in the workspace.

\paragraph{Implementation of S2I} For segment selection, we set the segment representation dimension to 256, use 64 neighbors for distance-weighted voting, and set the classification threshold $\delta_c$ to 0.5. We augment 500 positive and 500 negative samples per expert demonstration. In state-based simulation environments, we use empty images as starting and ending images in the representation model. The representation model is trained with an SGD optimizer at a learning rate of 0.005. We set the maximum angle threshold $\delta_\theta$ for trajectory optimization to 75$^\circ$.

\subsection{Simulation Results}\label{sec:sim}

\begin{table}[t]
\renewcommand{\arraystretch}{1.1}
    \centering
    \setlength\tabcolsep{1.4pt}
    \begin{tabular}{cc|crc}
    \toprule
    \begin{tabular}{c}\textbf{Selection} \\ \textbf{Level}\end{tabular} & \begin{tabular}{c} \textbf{w/wo Trajectory} \\ \textbf{Optimization}\end{tabular} & \multicolumn{1}{c}{\begin{tabular}{c}\textbf{Dataset}\\ \textbf{Utilization}\end{tabular}} & \begin{tabular}{c}\textbf{Average} \\ \textbf{Gain $\uparrow$} \end{tabular} & \begin{tabular}{c}\textbf{Average} \\ \textbf{Rank $\downarrow$} \end{tabular}\\ 
    \midrule
    -  & & 100.00\% & \color{darkgray}{$0.00\%$} & 4.00 \\
    -  & \checkmark & 100.00\% & \color{darkgreen}{$-27.48\%$} & 6.00\\
    \midrule
    Demonstration & & 70.93\% & \color{darkred}{$+0.81\%$} & 3.67\\
    Demonstration & \checkmark &  100.00\% & \color{darkred}{$+14.16\%$} & 2.33\\
    \midrule
    Segment & & 84.50\% & {\color{darkred}$+1.84\%$} & 3.00\\
    Segment &  \checkmark & 100.00\% & \color{darkred}{$\mathbf{+17.07\%}$} & \textbf{1.83}\\
    \bottomrule
    \end{tabular}
    \caption{\textbf{Ablation Results of Different S2I Design Choices.} BC-RNN average performance gains, as well as the average rank of the performance gain, on the state-based RoboMimic benchmark, are reported. The results are averaged over 6 experiments of 5 different seeds, based on the checkpoint with the best performance.}
    \label{tab:ablation}
    \vspace{-0.5cm}
\end{table}

\textbf{S2I can improve the performance of various downstream policies by optimizing the mixed-quality demonstration data (Q1).} Fig. \ref{fig:result}, Tab. \ref{tab:stability} and Tab.~\ref{tab:res-dp} illustrate that S2I consistently outperforms baseline methods in simulation environments. This underscores the effectiveness of S2I in handling such data. In line with \cite{dp}, we also report the average performance of the last 10 checkpoints to provide a representative assessment during training. S2I continues to surpass previous methods by a considerable margin ($\color{darkred}\sim +20\%$), demonstrating its ability to filter and optimize low-quality segments for substantial performance gains.


\footnotetext{ILEED~\cite{ileed} is only compatible with policies employing Gaussian mixture heads and is therefore excluded from DP experiments. PUBC~\cite{pubc} is designed for state inputs, so we exclude it in the image-based experiments.}

\textbf{All types of data, as long as carefully optimized, are useful for policy learning (Q2).} When naively discarding low-quality demonstrations, as in baselines like L2D and PUBC, policy performance drops significantly and the dataset utilization rates are relatively low, indicating that even suboptimal demonstrations are valuable for policy learning. \textit{These low-quality demonstrations can expand the exploration range in the state space.} Therefore, it is crucial to leverage the valuable insights they provide and \textit{fully utilize the mixed-quality dataset}. Thanks to our proposed trajectory optimization module, S2I offers an effective approach to optimize low-quality segments without altering the dataset's exploration range, resulting in improved policy performance.

\subsection{Ablations}\label{sec:ablation}

\textbf{Segment-level selection and trajectory optimization (with action relabeling) are both crucial for performance improvement (Q3).}
We ablate the selection level and the trajectory optimization module. For demonstration-level selection, we assess demonstration quality by averaging the quality results of each segment. The ablation results in Tab.~\ref{tab:ablation} show that without selection, performance drops when only trajectory optimization is applied ($\color{darkgreen}-27.48\%$), as high-quality segments are already optimal. Selecting at demonstration or segment level improves performance marginally ($\color{darkred}+0.81\%$ and $\color{darkred}+1.84\%$). Introducing trajectory optimization with action relabeling boosts both dataset utilization and performance, with segment-level application achieving the best results ($\color{darkred}+17.07\%$) by refining actions in low-quality data. These results validate S2I's design choices.

\begin{table}[t]
\renewcommand{\arraystretch}{1.1}
    \centering
    \setlength\tabcolsep{3.5pt}
    \begin{tabular}{c|c|cc|ccc}
    \toprule
    \multirow{2}{*}{\textbf{Method}} & \textbf{\textit{Tissue}} & \multicolumn{2}{c|}{\textbf{\textit{Cup}}} & \multicolumn{3}{c}{\textbf{\textit{Pen}}} \\
     & - & 1 cup & 2 cups & 1 pen & 2 pens & 3 pens \\
    \midrule
    ACT~\cite{act} & 0.00\% & 40.00\% & 35.00\% & 50.00\% & 25.00\% & 16.67\%  \\
    ACT + S2I & \textbf{25.00\%} & \textbf{55.00\%} & \textbf{45.00\%} & \textbf{60.00\%} & \textbf{40.00\%} & \textbf{23.33\%} \\ 
    \midrule
    DP~\cite{dp} & 10.00\% & 35.00\% & 30.00\% & 50.00\% & 30.00\% & 26.67\% \\
    DP + S2I & \textbf{65.00\%} & \textbf{55.00\%} & \textbf{50.00\%} & \textbf{70.00\%} & \textbf{50.00\%} & \textbf{36.67\%} \\
    \midrule
    RISE~\cite{rise}  & \textbf{90.00\%} & 90.00\% & 87.50\% & 70.00\% & 50.00\% & 40.00\% \\
    RISE + S2I & \textbf{90.00\%} & 
    \textbf{95.00\%} & \textbf{95.00\%} & \textbf{90.00\%} & \textbf{80.00\%} & \textbf{60.00\%} \\
    \bottomrule
    \end{tabular}
    \caption{\textbf{Results of Real-World Experiments.} The policy success (completion) rates over 20 trials for each setup are reported. }\label{tab:real-world}\vspace{-0.5cm}
\end{table}
\subsection{Real-World Results}\label{sec:real-world}

\textbf{S2I effectively optimizes real-world mixed-quality demonstration data, enhancing the performance of various downstream policies in real-world experiments (Q4).} The comprehensive results shown in Tab~\ref{tab:real-world} reveal that S2I can significantly improve performance for various downstream policies across different tasks, especially for those with initially low success rates. Hence, S2I is also consistently effective in utilizing real-world mixed-quality demonstrations and can substantially contribute to a better robotic manipulation policy.
\section{Conclusion}\label{sec:conclusion}

This paper introduces ``\textit{Select Segments to Imitate}'' (S2I), a framework designed to effectively use  mixed-quality demonstrations for robotic manipulation policy learning by focusing on segment-level selection and optimization. S2I extracts high-quality segments via contrastive learning and refines lower-quality ones through trajectory optimization and action relabeling, ensuring effective learning of corrective behaviors while fully utilizing the demonstration dataset. Extensive experiments in both simulated and real-world environments demonstrate the robust ability of S2I to optimize demonstrations and significantly boost robotic manipulation policy performance. While S2I shows promising results across various tasks and policies, it may struggle in handling robot trajectories with complex rotations. Future work includes addressing this issue, applying S2I to large-scale robotic datasets~\cite{rh20t}, and extending S2I to optimize language-guided demonstrations.

\section*{Acknowledgement}


This work was supported by the National Key Research and Development Project of China (No. 2022ZD0160102), the National Key Research and Development Project of China (No. 2021ZD0110704), Shanghai Artificial Intelligence Laboratory, XPLORER PRIZE grants.
 
\printbibliography

\clearpage
\section*{APPENDIX}

\subsection{Experiment Details}

\paragraph{Simulation} All experiments in the state-based RoboSuite simulation environment~\cite{robosuite} are conducted across seeds 1, 42, 233, 3407, and 9989, and the reported results are averaged across these seeds. To ensure a fair comparison in the simulation environment, we randomly generate 50 initialization setups for each seed and use the same generated setups across all methods during evaluations. Furthermore, demonstrations for each task are selected based on specific modular arithmetic rules to maintain a balanced distribution of demonstration qualities, as described in Appendix I.

\paragraph{Real-World} All experiments in the real-world robot platform are conducted with 20 consecutive trials, with the average success (completion) rate reported. The objects are randomly placed in the workspace. For trials with the same index across different methods, we keep the object positions in the workspace relatively fixed for fairness considerations.

\subsection{Baseline Implementations}

We introduce the details of baseline implementations here, including L2D~\cite{l2d}, ILEED~\cite{ileed}, and PUBC~\cite{pubc}. Moreover, we also conduct experiments of AWE~\cite{awe} to evaluate its performance in handling mixed-quality demonstration data.

\paragraph{L2D} We use the \textit{oracle} classifier as the replacement of the L2D classifier due to similar accuracy. Then, we keep the demonstrations with ``better'' and ``okay'' labels while discarding those with ``worse'' labels to form the dataset for downstream policy learning.

\paragraph{ILEED} Following the instructions in the original paper and utilizing \href{https://github.com/Stanford-ILIAD/ILEED}{the official code} as a reference, we implement two versions of ILEED: ILEED (\textit{w. labels}) and ILEED (\textit{wo. labels}). The former uses quality labels from the dataset as the expertise labels, and the latter assumes no quality labels are available and treats the expertise labels of all demonstrations as the same. The ILEED in Fig.~\ref{fig:result} and Tab.~\ref{tab:stability} correspond to the former one (\textit{w. labels}), which empirically demonstrates better performance. We use a state encoder (with softmax head) to extract information from the observation, which is combined with the expertise parameter $\omega$ as the weight. This process generates a scaling coefficient for the output variance of the Gaussian mixture head. The details of ILEED hyperparameters are shown in Tab.~\ref{tab:ileed-hyperparam}.

\begin{table}[h]
    \centering
    \begin{tabular}{c|c}
         \toprule
         \textbf{Hyperparameters} & \textbf{\textit{Lift / Can / Square}} \\
         \midrule
         State Encoder & 3–layer MLP + Softmax  \\
         Activation & ReLU \\
         Hidden Size & 16  \\
         Iterations & 2000  \\
         Learning Rate of $\omega$ & 1e-4 \\
         State Embedding Dimension & 2  \\
         \bottomrule
    \end{tabular}
    \caption{\textbf{Hyperparameters of ILEED~\cite{ileed} in the Experiments.}}
    \label{tab:ileed-hyperparam}
\end{table}

\paragraph{PUBC} We slightly modify \href{https://github.com/RedmondLabUCD/Positive-Unlabeled-Behavioural-Cloning}{the official code} to generate positive dataset iteratively from the mixed-quality dataset $\mathcal{D}$ and seed positive dataset $\mathcal{D}^*$. The generated positive dataset is used for downstream policy learning. Because PUBC is designed specifically for 1D state inputs, we only conduct evaluations in state-based RoboMimic environments. We have carefully tuned the hyperparameters of PUBC and empirically determined the hyperparameters for our experiments based on their performances. The details of PUBC hyperparameters are shown in Tab.~\ref{tab:pubc-hyperparam}. 

\begin{table}[h]
    \centering
    \begin{tabular}{c|ccc}
         \toprule
         \textbf{Hyperparameters} & \textbf{\textit{Lift}} & \textbf{\textit{Can}} & \textbf{\textit{Square}} \\
         \midrule
         Models in Ensemble & 3 & 3 & 3 \\
         Iterations & 10 & 10 & 10 \\
         Negative Sampler & full & full & full \\
         Epochs per Iteration & 20 & 20 & 20 \\
         Batch Size & 1024 & 1024 & 1024 \\
         Threshold Fitting Pow & 8 & 10 & 10 \\
         Threshold High Bound & 0.5 & 0.5 & 0.5 \\
         \bottomrule
    \end{tabular}
    \caption{\textbf{Hyperparameters of PUBC~\cite{pubc} in the Experiments.}}
    \label{tab:pubc-hyperparam}
\end{table}

\paragraph{AWE} We directly use \href{https://github.com/lucys0/awe/}{the official code} to optimize all mixed-quality demonstration trajectories. All hyperparameters remain the same.

\subsection{Robotic Manipulation Policy Implementations}

Since S2I does not require joint design or joint training with the policy, we do not need to modify the hyperparameters of the policies. Therefore, we inherit the original hyperparameters in their papers~\cite{dp, robomimic, rise, act}. In simulation environments, for BC-RNN (relative action space), we set the training epochs as 2000 and save a checkpoint every 100 epochs, with 250 warm-up epochs; for state-based DP (absolute action space), we set the training epochs as 2000 and save a checkpoint every 100 epochs; for image-based DP, we set the training epochs as 900 and save a checkpoint every 50 epochs. For real-world experiments, we set the training epochs as 1000 and use the last checkpoint in evaluations for all policies. 

In terms of action representations, BC-RNN utilizes \textit{relative} action representations, while the rest methods all use \textit{absolute} action representations, with DP and RISE leveraging continuous 6D representation for rotation~\cite{rot6d}.

\subsection{Segment Representation Results}

\begin{figure}[b]
\vspace{-0.2cm}
    \centering
    \includegraphics[width=1\linewidth]{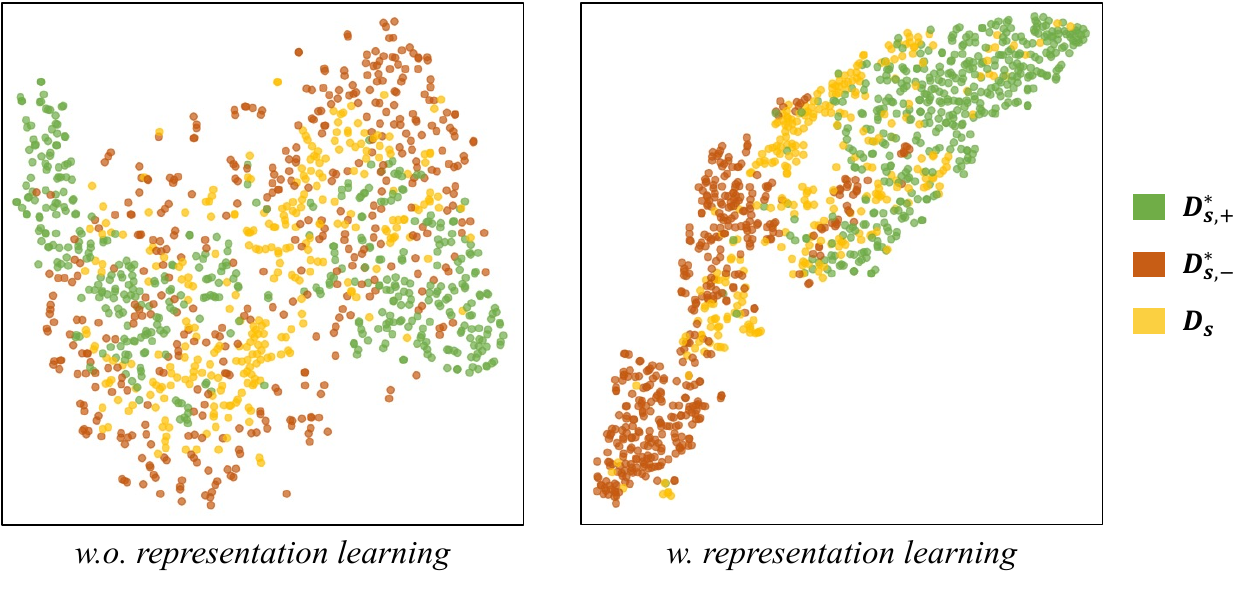}
    \caption{\textbf{$t$-SNE Visualizations of the Feature Space of \textit{Can}-100 \textit{w/wo} Segment Representation Learning.}}
    \label{fig:t-sne}\vspace{-0.4cm}
\end{figure}

\begin{table*}
\renewcommand{\arraystretch}{1.1}
    \centering
    \setlength\tabcolsep{4.5pt}
    \begin{tabular}{c|cc|cc|cc|rcl}
    \toprule
    \multirow{2}{*}{\textbf{Method}} & \multicolumn{2}{c|}{\textbf{\textit{Lift}}} & \multicolumn{2}{c|}{\textbf{\textit{Can}}} & \multicolumn{2}{c|}{\textbf{\textit{Square}}} & \multicolumn{3}{c}{\multirow{2}{*}{\begin{tabular}{c}\textbf{Average Gain $\uparrow$}\\ (best / last 10 average)\end{tabular}}} \\ 
    & 10 & 30 & 30 & 100 & 50 & 150\\
    \midrule
    Original & 0.520 / 0.372  & 0.868 / 0.769 & 0.460 / 0.357 & 0.900 / \textbf{0.829} & 0.260 / 0.177 & 0.652 / 0.511  & $\color{darkgray}0.00\%$ &/& $\color{darkgray}0.00\%$\\ 
    \midrule
    L2D~\cite{l2d} (\textit{Oracle}) & 0.388 / 0.330 & 0.836 / 0.738 & 0.364 / 0.278 & 0.880 / 0.792 & 0.244 / 0.134 & 0.556 / 0.438 & $\color{darkgreen}-12.17\%$ &/& $\color{darkgreen}-12.37\%$\\
    PUBC~\cite{pubc} & 0.396 / 0.300 & 0.464 / 0.365 & 0.460 / 0.361 & 0.860 / 0.760 & 0.188 / 0.123 & 0.624 / 0.503  & $\color{darkgreen}-18.31\%$ &/& $\color{darkgreen}-18.58\%$ \\
    ILEED~\cite{ileed} (\textit{w. labels}) & 0.520 / 0.432 & 0.860 / 0.776 & 0.484 / 0.374 & 0.896 / 0.793 & 0.244 / 0.171 & 0.592 / 0.469 & $\color{darkgreen}-1.92\%$ &/& $\color{darkred}+0.97\%$ \\ 
    ILEED~\cite{ileed} (\textit{wo. labels}) & \textbf{0.548} / \textbf{0.446} & 0.860 / 0.740 & 0.284 / 0.213 & 0.856 / 0.734 & 0.236 / 0.154 & 0.572 / 0.471 & $\color{darkgreen}-10.03\%$ &/& $\color{darkgreen}-9.36\%$\\
    \midrule
    S2I-ILEED & 0.544 / 0.436 & 0.880 / 0.768 & 0.568 / 0.454 & 0.912 / 0.822 & 0.236 / 0.174 & 0.608 / 0.482 & $\color{darkred}+2.47\%$ &/& $\color{darkred}+5.97\%$  \\
    S2I (\textit{ours}) & 0.532 / 0.417 & \textbf{0.972} / \textbf{0.898} & \textbf{0.672} / \textbf{0.556} & \textbf{0.916} / 0.826 & \textbf{0.336} / \textbf{0.228} & \textbf{0.724} / \textbf{0.581} & $\color{darkred}\mathbf{+17.07\%}$ &/& $\color{darkred}\mathbf{+21.16\%}$ \\
    \bottomrule
    
    \end{tabular}
    \caption{\textbf{Detailed BC-RNN Performance (Success Rate) on the State-Based RoboMimic Benchmark.} This table includes the full results of Fig.~\ref{fig:result} and Tab.~\ref{tab:stability}. Following \cite{dp}, the same format of (best performance) / (average of last 10 checkpoints) is utilized here. The results are averaged across 5 different seeds and 50 different initial environment setups (250 in total). }\label{tab:simulation-all}
   \vspace{-0.2cm}
\end{table*}

\begin{table*}
\renewcommand{\arraystretch}{1.1}
    \setlength\tabcolsep{4.5pt}
    \centering
    \begin{tabular}{cc|ccc|rcl|rcl}
    \toprule
    \textbf{Environment} & \textbf{Method} & \textbf{\textit{Lift}}-10 & \textbf{\textit{Can}}-30 & \textbf{\textit{Square}}-50 & \multicolumn{3}{c|}{\textbf{Average Gain $\uparrow$}} & \multicolumn{3}{c}{\textbf{Worst Gain $\uparrow$}} \\ 
    \midrule
    \multirow{5}{*}{\textbf{State-based}} & Original &  0.952 / \textbf{0.912}  & 0.824 / 0.749  & 0.504 / 0.409 & $\color{darkgray}0.00\%$ &/& $\color{darkgray}0.00\%$ & $\color{darkgray}0.00\%$ &/& $\color{darkgray}0.00\%$ \\
    \cmidrule{2-11}
    & L2D~\cite{l2d} (\textit{Oracle}) & 0.544 / 0.457  & 0.796 / 0.734  & 0.368 / 0.258 & $\color{darkgreen}-24.41\%$ &/& $\color{darkgreen}-29.60\%$ & $\color{darkgreen}-42.86\%$ &/& $\color{darkgreen}-49.89\%$  \\
    & PUBC~\cite{pubc} & 0.792 / 0.723 & 0.756 / 0.680 & 0.316 / 0.250 & $\color{darkgreen}-20.79\%$ &/& $\color{darkgreen}-22.94\%$ & $\color{darkgreen}-37.30\%$ &/& $\color{darkgreen}-38.88\%$  \\
    & AWE~\cite{awe} & 0.688 / 0.605 & 0.836 / 0.774  & 0.552 / 0.482 & $\color{darkgreen}-5.58\%$ &/& $\color{darkgreen}-4.16\%$ & $\color{darkgreen}-27.73\%$ &/& $\color{darkgreen}-33.66\%$  \\ 
    \cmidrule{2-11}
    & S2I (\textit{ours}) & \textbf{0.964} / 0.906 & \textbf{0.840} / \textbf{0.785} & \textbf{0.580} / \textbf{0.500} & $\color{darkred}\mathbf{+6.09\%}$ &/& $\color{darkred}\mathbf{+8.80\%}$& $\color{darkred}\mathbf{+1.26\%}$ &/& $\color{darkgreen}\mathbf{-0.66\%}$  \\
    \bottomrule
    \toprule
    \multirow{4}{*}{\textbf{Image-based}} & Original &  0.964 / 0.743  & 0.640 / 0.547 & 0.568 / 0.492 & $\color{darkgray}0.00\%$ &/& $\color{darkgray}0.00\%$ & $\color{darkgray}0.00\%$ &/& $\color{darkgray}0.00\%$   \\
    \cmidrule{2-11}
    & L2D~\cite{l2d} (\textit{Oracle})& 0.708 / 0.442 & 0.524 / 0.439 & 0.444 / 0.377 & $\color{darkgreen}-22.17\%$ &/& $\color{darkgreen}-27.88\%$  & $\color{darkgreen}-26.56\%$ &/& $\color{darkgreen}-40.51\%$ \\
    & AWE~\cite{awe} &  \textbf{0.996} / \textbf{0.951} & \textbf{0.728} / \textbf{0.660} & 0.464 / 0.385 & $\color{darkred}\mathbf{+8.97\%}$ &/& $\color{darkgreen}-0.41\%$ & $\color{darkgreen}-18.31\%$ &/& $\color{darkgreen}-21.75\%$\\ 
    \cmidrule{2-11}
    & S2I (\textit{ours}) & 0.956 / 0.867 & 0.712 / 0.624 & \textbf{0.604} / \textbf{0.506} & $\color{darkred}+5.59\%$ &/& $\color{darkred}\mathbf{+11.20\%}$ & $\color{darkgreen}\mathbf{-0.83\%}$ &/& $\color{darkred}\mathbf{+2.85\%}$ \\ 
    \bottomrule
    \end{tabular}
    \caption{\textbf{Detailed DP Performance (Success Rate, with additional experiments of AWE~\cite{awe}) on the RoboMimic Benchmark.} The success rates are averaged across 5 different seeds. Following \cite{dp}, the same format of (best performance) / (average of last 10 checkpoints) is reported here. }\label{tab:res-dp}\vspace{-0.4cm}
\end{table*}

We use $t$-SNE to visualize the segment representation of \textbf{\textit{Can}}-30 in Fig.~\ref{fig:t-sne}, to illustrate how the contrastive learning process clusters similar quality samples in the feature space. This clustering organizes the segment data meaningfully and effectively helps distance-weighted voting by distinguishing different quality types.

\subsection{Additional Simulation Results}
We provide the detailed RoboMimic state-based results in Tab.~\ref{tab:simulation-all}, including methods ILEED (\textit{w. labels}) and S2I-ILEED. In the latter one, we assume that two demonstrators have different expertise: one \textit{proficient} producing all high-quality demonstrations and one \textit{rusty} producing all low-quality demonstrations. We classify demonstrations into high-quality and low-quality sets using the selection results from S2I, averaged across all segments within each demonstration. Demonstrations in the high-quality set are attributed to the \textit{proficient} demonstrator, while those in the low-quality set are attributed to the \textit{rusty} demonstrator.

\textbf{Using ground-truth expertise labels of demonstrators in the dataset can improve the performance of ILEED.} If we treat all demonstrations as having the same level of expertise, performance is inferior compared to using ground-truth expertise labels. This shows that ground-truth expertise labels in the dataset can reflect the quality of the demonstrations to some extent. However, if the expertise labels are unavailable, vanilla ILEED may yield unsatisfactory results.

\textbf{Using the quality labels of the demonstrations determined by S2I can further improve the performance of ILEED.} We observe that demonstrators with \textit{okay} expertise labels can sometimes produce low-quality demonstrations and those with \textit{worse} labels can occasionally produce high-quality demonstrations (Fig.~\ref{fig:quality}). This variability can cause inconsistencies in the demonstration sets attributed to each demonstrator. In S2I-ILEED, by assuming one \textit{proficient} and one \textit{rusty} demonstrator and classifying demonstrations into two sets based on these \textit{virtual} demonstrators, we achieve greater consistency in quality within each set. Consequently, the expertise labels derived in this manner are more consistent and effective than the ground-truth labels provided in the dataset. Experimental results confirm this, showing that S2I-ILEED further enhances the policy performance with ILEED (${\color{darkgreen}-1.92\%} \rightarrow {\color{darkred}+2.47\%}$ / ${\color{darkred}+0.97\%} {\rightarrow \color{darkred}+5.97\%}$).

\begin{table*}
\renewcommand{\arraystretch}{1.1}
    \centering
    \setlength\tabcolsep{1pt}
    \begin{tabular}{ccc|rrrrrr|crc}
    \toprule
    \multirow{3}{*}{\begin{tabular}{c}\textbf{Selection} \\ \textbf{Level}\end{tabular}} & \multirow{3}{*}{\begin{tabular}{c}\textbf{w/wo Trajectory} \\ \textbf{Optimization}\end{tabular}} & \multirow{3}{*}{\begin{tabular}{c} \textbf{w/wo Action} \\ \textbf{Relabeling}\end{tabular}} & \multicolumn{6}{c|}{\textbf{Performance Gain} $\uparrow$} & \multirow{3}{*}{\begin{tabular}{c}\textbf{Dataset}\\ \textbf{Utilization}\end{tabular}} & \multirow{3}{*}{\begin{tabular}{c}\textbf{Average} \\ \textbf{Gain $\uparrow$} \end{tabular}} & \multirow{3}{*}{\begin{tabular}{c}\textbf{\# Tasks with} \\ \textbf{Improvements $\uparrow$} \end{tabular}}\\ 
    & & & \multicolumn{2}{c}{\textbf{\textit{Lift}}} & \multicolumn{2}{c}{\textbf{\textit{Can}}} & \multicolumn{2}{c|}{\textbf{\textit{Square}}} & & \\
    & & & \multicolumn{1}{c}{10} & \multicolumn{1}{c}{30} & \multicolumn{1}{c}{30} & \multicolumn{1}{c}{100} & \multicolumn{1}{c}{50} & \multicolumn{1}{c|}{150} & & \\ 
    \midrule
    - & & & {\color{darkgray}$0.00\%$} & {\color{darkgray}$0.00\%$}  & {\color{darkgray}$0.00\%$}  & {\color{darkgray}$0.00\%$}  & {\color{darkgray}$0.00\%$}  & {\color{darkgray}$0.00\%$}  & 100.00\% & {\color{darkgray}$0.00\%$} & 0 / 6\\
    - & \checkmark & & {\color{darkgreen}$-51.54\%$} & {\color{darkgreen}$-27.19\%$} & {\color{darkgreen}$-46.09\%$} & {\color{darkgreen}$-34.67\%$} & {\color{darkgreen}$-50.77\%$} & {\color{darkgreen}$-40.49\%$} & 70.30\% & {\color{darkgreen}$-41.79\%$} & 0 / 6 \\
    -  & \checkmark & \checkmark & {\color{darkgreen}$-25.38\%$} & {\color{darkgreen}$-3.69\%$} & {\color{darkgreen}$-1.74\%$} & {\color{darkgreen}$-11.56\%$} & {\color{darkgreen}$-47.69\%$} & {\color{darkgreen}$-74.85\%$} & 100.00\% & {\color{darkgreen}$-27.48\%$} & 0 / 6 \\
    \midrule
    Demonstration & & & {\color{darkred}$+26.15\%$} & {\color{darkgreen}$-0.46\%$} & {\color{darkgreen}$-9.57\%$} & {\color{darkred}$+2.22\%$} & {\color{darkgray}$0.00\%$} & \color{darkgreen}{$-13.50\%$} & 70.93\% & \color{darkred}{$+0.81\%$} & 2 / 6 \\
    Demonstration & \checkmark & & {\color{darkgreen}$-5.38\%$}  & {\color{darkred}$+5.53\%$}  & {\color{darkred}$+13.04\%$}  & {\color{darkred}$+4.44\%$}  & {\color{darkgreen}$-32.31\%$}  & {\color{darkred}$+4.91\%$}  & 87.05\% & {\color{darkgreen}$-1.63\%$} & 4 / 6 \\
    Demonstration & \checkmark & \checkmark & {\color{darkgreen}$-6.15\%$}  & {\color{darkred}$+13.36\%$}  & {\color{darkred}$+35.65\%$}  & {\color{darkred}$+0.89\%$}  & {\color{darkred}$+33.85\%$}  & {\color{darkred}$+7.36\%$}  & 100.00\% & {\color{darkred}$+14.16\%$} & 5 / 6 \\
    \midrule
    Segment & & & {\color{darkgreen}$-7.69\%$} & {\color{darkred}$+0.46\%$} & {\color{darkred}$+6.09\%$} & {\color{darkred}$+2.67\%$} & {\color{darkred}$+7.69\%$} & {\color{darkred}$+1.84\%$} & 84.50\% &{\color{darkred}$+1.84\%$} & 5 / 6 \\
    Segment & \checkmark & & {\color{darkred}$+3.08\%$}  & {\color{darkgreen}$-7.83\%$}  & {\color{darkred}$+46.96\%$}  & {\color{darkred}$+0.89\%$}  & {\color{darkgreen}$-1.54\%$}  & {\color{darkred}$+0.61\%$}  & 91.04\% &  {\color{darkred}$+7.03\%$} & 4 / 6 \\
    Segment & \checkmark & \checkmark & {\color{darkred}$+2.31\%$}  & {\color{darkred}$+11.98\%$}  & {\color{darkred}$+46.09\%$}  & {\color{darkred}$+1.78\%$}  & {\color{darkred}$+29.23\%$}  & {\color{darkred}$+11.04\%$}  & 100.00\% & {\color{darkred}$\mathbf{+17.07\%}$} & \textbf{6 / 6} \\
    \bottomrule
    \end{tabular}
    \caption{\textbf{Detailed Ablation Results of Different S2I Design Choices}. BC-RNN performance gains on the state-based RoboMimic benchmark are reported. We also count the number of tasks with improvements as a consistency metric. The results are averaged over 6 experiments of 5 different seeds, based on the checkpoint with the best performance.}\label{tab:ablation-all}
    \vspace{-0.2cm}
\end{table*}

\begin{table*}
\renewcommand{\arraystretch}{1.1}
    \centering
    \setlength\tabcolsep{4pt}
    \begin{tabular}{c|c|c|cc|cc|ccc|ccc}
    \toprule
    \multirow{3}{*}{\textbf{Method}} & \multicolumn{2}{c|}{\textbf{\textit{Tissue}}} & \multicolumn{4}{c|}{\textbf{\textit{Cup}}} & \multicolumn{6}{c}{\textbf{\textit{Pen}}} \\
     & 20\%-80\% & 80\%-20\% & \multicolumn{2}{c|}{20\%-80\%} & \multicolumn{2}{c|}{80\%-20\%} & \multicolumn{3}{c|}{20\%-80\%} & \multicolumn{3}{c}{80\%-20\%} \\
     & - & - & 1 cup & 2 cups & 1 cup & 2 cups & 1 pen & 2 pens & 3 pens & 1 pen & 2 pens & 3 pens \\
    \midrule
    ACT~\cite{act} & 0.00\% & 5.00\% & 30.00\% & 22.50\% & 45.00\% & 35.00\% & 40.00\% & 25.00\% & 20.00\% & 50.00\% & 30.00\% & 26.67\% \\
    ACT~\cite{act} + S2I & \textbf{15.00\%} & \textbf{20.00\%} & \textbf{45.00\%} & \textbf{30.00\%} & \textbf{60.00\%} & \textbf{37.50\% }& \textbf{50.00\%} & \textbf{35.00\%} & \textbf{23.33\%} & \textbf{60.00\%} & \textbf{45.00\%} & \textbf{36.67\%}\\ 
    \midrule
    DP~\cite{dp} & 25.00\% & 35.00\% & 30.00\% & 25.00\% & 50.00\% & 40.00\% & 40.00\% & 30.00\% & 13.33\% & \textbf{70.00\%} & 40.00\% & 33.33\% \\
    DP~\cite{dp} + S2I & \textbf{85.00\%} & \textbf{60.00\%} & \textbf{55.00\%} & \textbf{45.00\%} & \textbf{55.00\%} & \textbf{45.00\%} & \textbf{70.00\%} & \textbf{45.00\%} & \textbf{30.00\%} & \textbf{70.00\%} & \textbf{45.00\%} & \textbf{36.67\%} \\
    \midrule
    RISE~\cite{rise} & 90.00\% & 85.00\% & 80.00\% & 72.50\% & \textbf{95.00\%} & \textbf{95.00\%} & 65.00\% & 42.50\% & 35.00\% & 85.00\% & 72.50\% & 63.33\% \\
    RISE~\cite{rise} + S2I & \textbf{95.00\%} & \textbf{95.00\%} & \textbf{90.00\%} & \textbf{80.00\%} & \textbf{95.00\%} & \textbf{95.00\%} & \textbf{80.00\%} & \textbf{70.00\%} & \textbf{51.67\%} & \textbf{95.00\%} & \textbf{87.50\%} & \textbf{73.33\%} \\
    \bottomrule
    \end{tabular}
    \caption{\textbf{Additional Results of Real-World Experiments.} This table reports the policy success (completion) rate over 20 trials for each task and data mixture setups (\textbf{20\%-80\%}: 20\% expert demonstrations and 80\% suboptimal demonstrations; \textbf{80\%-20\%}: 80\% expert demonstrations and 20\% suboptimal demonstrations). }\label{tab:real-world-additional}\vspace{-0.4cm}
\end{table*}

\textbf{S2I also achieves consistent performance with DP on both state-based and image-based RoboMimic benchmarks.} The detailed DP performance results on the RoboMimic benchmark are shown in Tab.~\ref{tab:res-dp}. We observe that S2I achieves consistent performance on both benchmarks (with an average performance gain of $\color{darkred}+6.09\%$ / $\color{darkred}+8.80\%$ and $\color{darkred}+5.59\%$ / $\color{darkred}+11.20\%$ for state-based and image-based benchmarks respectively), outperforming prior baselines (include AWE~\cite{awe}, which we will discuss later) by a large margin, demonstrating the effectiveness of S2I when combining with different downstream robot manipulation policies.

\textbf{AWE, while achieves competitive performance with image-based DP on pick-and-place tasks like \textit{Lift} and \textit{Can}, fails in state-based environments like \textit{Lift} and complex fine-grained tasks like \textit{Square}. }  We suspect that this inconsistency can be attributed to the fact that AWE does not adequately preserve outliers in the state-based domain, which is crucial for handling the intricacies of such environments. A possible explanation is that the image domain may be less sensitive to the discrepancies in state representations, allowing the model to overlook critical details without severely impacting performance. In contrast, state-based environments are more prone to overfitting incorrect behaviors, which significantly hampers performance in benchmarks designed for these settings (such as state-based \textbf{\textit{Lift}}, AWE results in $\color{darkgreen}{-33.66\%}$ performance degradation). Another example is the image-based \textbf{\textit{Square}}, the fine-grained tasks might become more difficult with the possible preserved outliers after AWE optimization, leading to a $\color{darkgreen}{-21.75\%}$ performance drop. Therefore, AWE is better at optimizing proficient data but struggles with edge cases, limiting its generalization. On the contrary, S2I maintains consistent performance for different tasks in both state-based and image-based environments (with the worse performance gain of $\color{darkgreen}-0.66\%$ and $\color{darkgreen}-0.83\%$ for state-based and image-based benchmark respectively), making the downstream policy more robust to the mixed-quality demonstration data.
\subsection{Additional Ablations}

We provide the full ablation results in Tab.~\ref{tab:ablation-all}, including the ablation of the action relabeling module. For demonstration-level selection, we assess demonstration quality by averaging the quality results of each segment instead of using the operator expertise labels provided in the dataset.

\textbf{Demonstration-level and segment-level selections offer limited performance improvement without optimizing low-quality data.} When simply discarding the low-quality data, demonstration-level and segment-level selections achieve average performance gains of $\color{darkred}+0.81\%$ and $\color{darkred}+1.84\%$, respectively.

\textbf{Segment-level selection outperforms demonstration-level selection in both average performance gain and improvement consistency.} Every variant with segment-level selection achieves better performance gain compared to its corresponding variant with demonstration-level selection. Moreover, from the number of tasks with improvements, we can observe that segment-level selection has more consistent performance improvements across various tasks than demonstration-level selection.

\textbf{Trajectory optimization (without action relabeling) on low-quality data improves performance while optimizing high-quality data might hurt performance.} We found that vanilla trajectory optimization yields better performance (${\color{darkred}+1.84\%} \rightarrow {\color{darkred}+7.03\%}$) after segment-level selection. However, applying trajectory optimization to all demonstrations (no selection) or low-quality demonstrations (demonstration-level selection) can degrade policy performance (${\color{darkgray}0.00\%} \rightarrow {\color{darkgreen}-41.79\%}$ and ${\color{darkred}+0.81\%} \rightarrow {\color{darkgreen}-1.63\%}$). We believe this occurs because trajectory optimization estimates a relatively optimal path, which enhances performance when optimizing low-quality segments but can harm performance with high-quality segments. Both scenarios (no selection and demonstration-level selection)  involve optimizing high-quality segments, leading to negative effects on the policy. The phenomenon also underscores the importance of segment-level selection for fine-grained quality assessment and filtering of the demonstration data.

\textbf{Action relabeling enhances the dataset utilizations and thus improves performances.} As stated in Sec.~\ref{sec:ablation}, low-quality demonstrations can expand the exploration range in the state space. Action relabeling enables the policy to fully utilize the dataset, offering a broader exploration range compared to methods that discard samples during selection or optimization.

\textbf{Our S2I framework, with segment-level selection, trajectory optimization, and action relabeling, yields the best performance improvement.} Our S2I framework achieves an average performance gain of $\color{darkred}+17.07\%$, demonstrating effectiveness in handling mixed-quality demonstrations.

\subsection{Additional Real-World Results}

We additionally collected 15 expert and 15 suboptimal demonstrations for each task, resulting in an expert dataset and a suboptimal dataset of 40 demonstrations each. We then create mixed-quality datasets by selecting demonstrations from both datasets in proportions of 20\%-80\%, 50\%-50\%, and 80\%-20\%, resulting in datasets with 50 demonstrations. Tab.~\ref{tab:real-world} has reported the results of a 50\%-50\% data mixture. We also provide results with other data mixtures in Tab.~\ref{tab:real-world-additional}.

\textbf{S2I demonstrates consistent effectiveness across varying proportions of mixed-quality data.}
Every method incorporating S2I shows better performance gains than its variant without S2I, demonstrating the robustness and versatility of S2I across tasks with mixed-quality demonstration data. This also includes more complex scenarios such as handling multiple cups or multiple pens, where S2I consistently enhances performance and proves to be a reliable improvement in various experimental settings.

\textbf{Methods with higher baseline performance show small gains from S2I.}
We observed that RISE already performs exceptionally well on the \textbf{\textit{Tissue}} task despite being trained on mixed-quality demonstrations, resulting in only marginal improvements (90.00\% $\rightarrow$ 95.00\% / 85.00\% $\rightarrow$ 95.00\%) after incorporating S2I. This suggests that RISE is inherently robust, and while S2I provides an additional boost, the gains are relatively modest. We believe this occurs because S2I is beneficial when there is more room for improvement, and in the case of RISE for easy tasks, its high baseline limits the potential for performance enhancement.

\textbf{Mixed data with higher proportions of low-quality samples benefits more from S2I.}
S2I proves more effective when dealing with mixed data containing higher proportions of low-quality samples. Generally, the 20\%-80\% data mixture (with 80\% low-quality data) benefits more from S2I than the 80\%-20\% data mixture.

\subsection{Qualitative Analyses of Demonstrations}

\begin{figure*}
    \centering
    \includegraphics[width=0.95\linewidth]{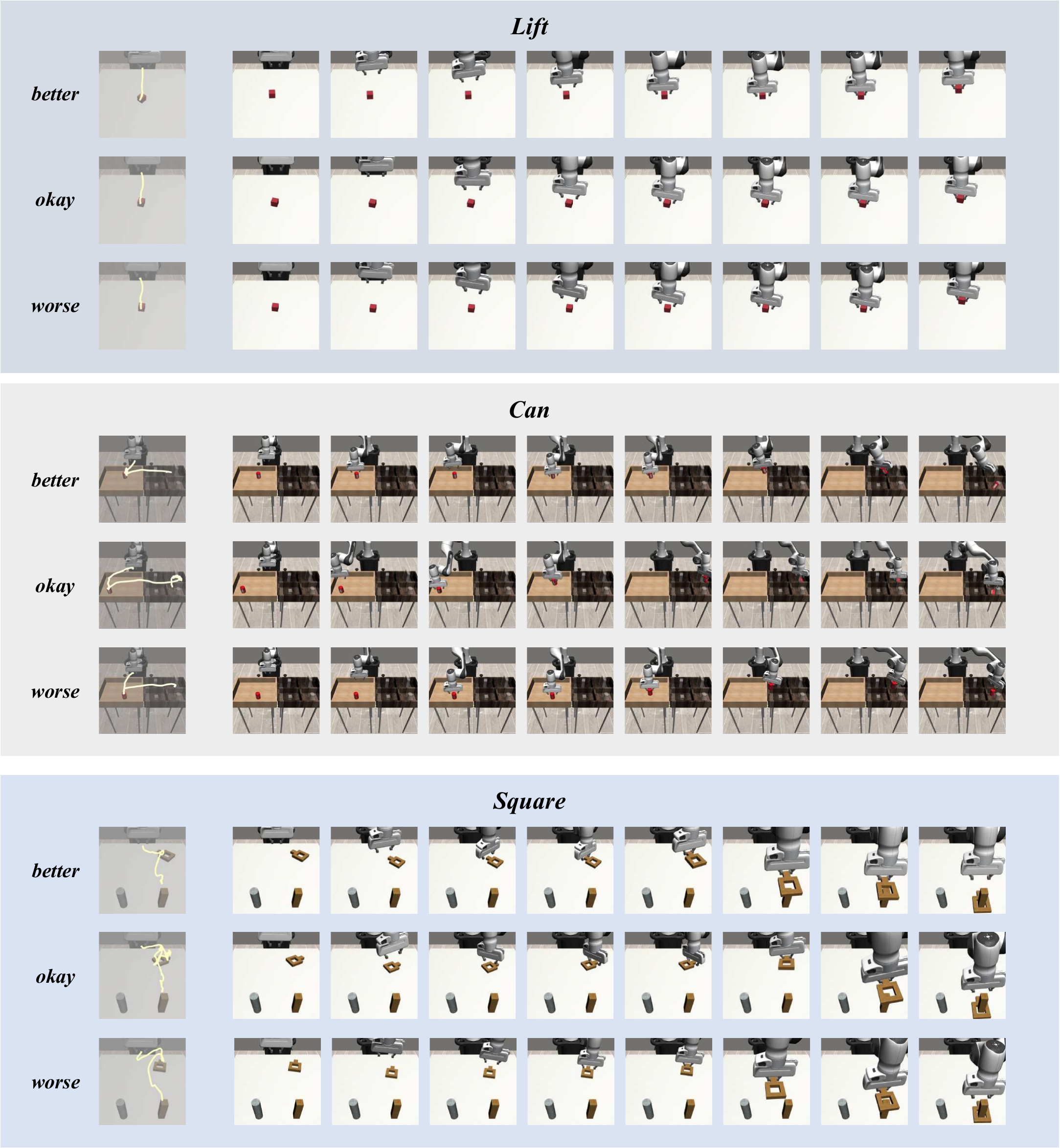}
    \caption{\textbf{Visualization of Demonstrations with Different Expertise Levels in the RoboMimic Multi-Human (MH) Dataset.} The RoboMimic dataset~\cite{robomimic} divides demonstrators into three levels of expertise: \textit{better}, \textit{okay}, \textit{worse}, and labels the demonstrations with the expertise of its demonstrator. However, we found that the \textit{better}-level demonstrators do not necessarily produce high-quality demonstrations, and \textit{worse}-level demonstrators may produce high-quality demonstrations, as visualized here.}
    \label{fig:quality} \vspace{-0.4cm}
\end{figure*}

\textbf{It is difficult to determine the quality of demonstrations simply from the expertise of the demonstrator.} We visualize the demonstrations in RoboMimic multi-human (MH) dataset~\cite{robomimic} in Fig.~\ref{fig:quality}, where we manually select a demonstration from the \textit{okay}-level, \textit{better}-level and \textit{worse}-level demonstrators respectively. It can be seen that the selected demonstrations from the \textit{worse}-level demonstrators are comparable to the selected demonstration from the \textit{better}-level demonstrators, suggesting that the expertise level of the demonstrator is not always a reliable indicator of demonstration quality. At times, even less experienced demonstrators can produce a small number of high-quality demonstrations, and even proficient demonstrators can make mistakes occasionally. 

\begin{figure*}
    \centering
    \includegraphics[width=0.95\linewidth]{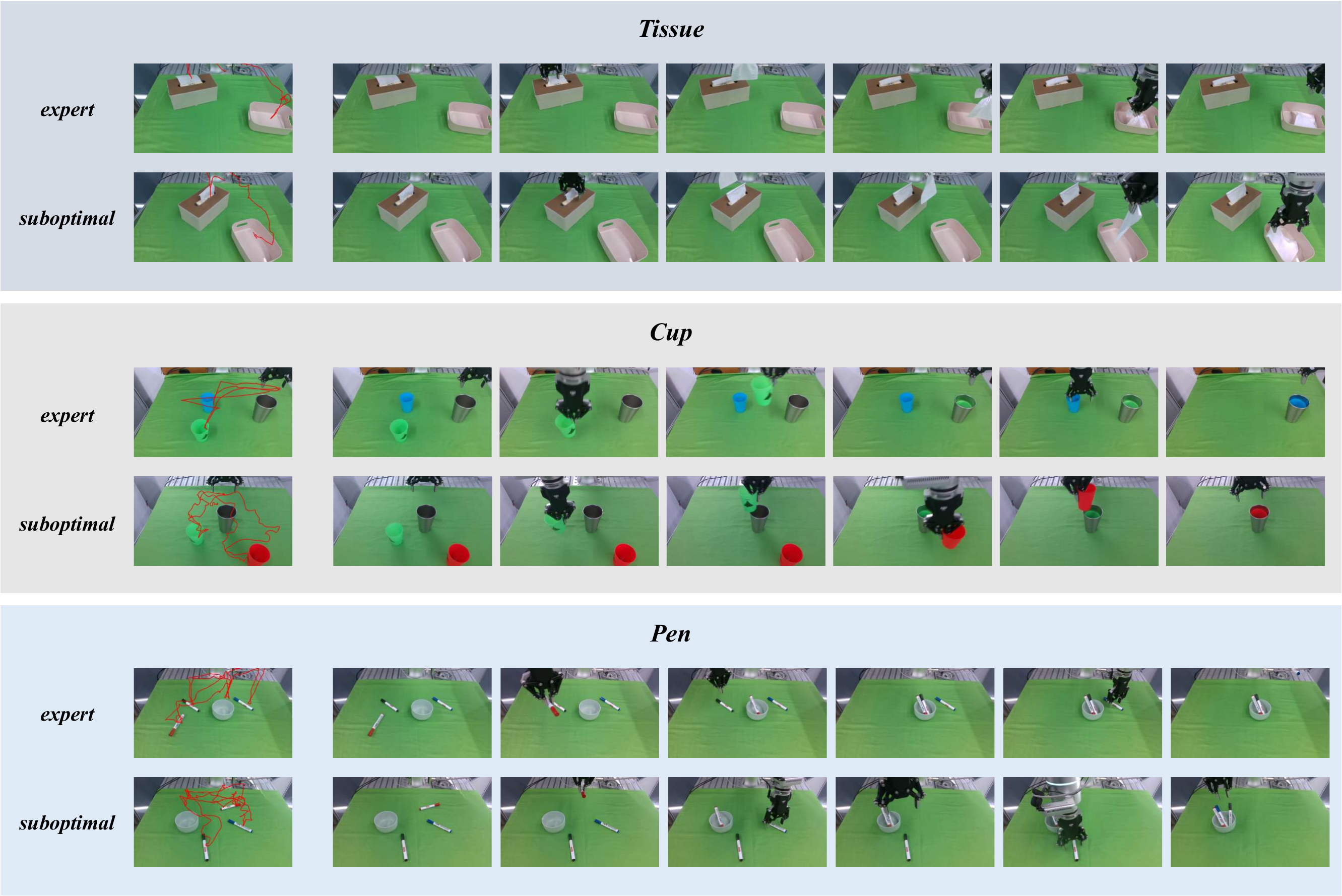}
    \caption{\textbf{Visualization of Demonstrations with Different Expertise Level in Real World.} }
    \label{fig:quality-rw} \vspace{-0.4cm}
\end{figure*}

Fig.~\ref{fig:quality-rw} shows the real-world mixed-quality demonstrations collected for our real-world experiments, revealing that expert demonstrations yield smooth trajectories, while suboptimal ones are usually much noisier. This noise complicates the downstream policy in effectively learning robot actions from the demonstrations and also validates our contrastive learning design of augmentation negative segments (\S \ref{sec:segment_selection}).

\subsection{Demonstration Selection Rules}\label{demo_select}

In our simulation environment, we selected multiple demonstrations for each task to maintain a balanced demonstration quality. The demonstration IDs for each task were chosen based on the following criteria (Tab.~\ref{tab:selection}), ensuring a diverse range of quality levels while keeping randomness.

\begin{table}[h]
\renewcommand{\arraystretch}{1.1}
    \centering
    \begin{tabular}{ccccc}
         \toprule
        \multirow{2}{*}{\textbf{Task}}   & \multirow{2}{*}{\textbf{Selection Rule}}  & \multicolumn{3}{c}{\textbf{\# Demonstrations}}\\ \cmidrule(lr){3-5}
        & & \textbf{\textit{Better}} & \textbf{\textit{Okay}} & \textbf{\textit{Worse}} \\
         \midrule
        \textbf{\textit{Lift}}-10    & ID mod 30 = 2 & 4 & 2 & 4 \\ 
        \textbf{\textit{Lift}}-30    & ID mod 10 = 1 & 10 & 10 & 10  \\
        \textbf{\textit{Can}}-30     & ID mod 10 = 1 & 10 & 10 & 10 \\
        \textbf{\textit{Can}}-100    & ID mod 3 = 1  & 34 & 34 & 32\\ 
        \textbf{\textit{Square}}-50  & ID mod 6 = 1  & 18 & 16 & 16 \\ 
        \textbf{\textit{Square}}-150 & ID mod 2 = 0  & 50 & 50 & 50 \\  
         \bottomrule
    \end{tabular}
\caption{\textbf{Demostration Selection Rules for Each Task.} This table describes the selection criteria for demonstration IDs based on their modular arithmetic rules for each task. }\label{tab:selection}
\end{table}

\end{document}